\documentclass[preprint]{elsarticle}

\usepackage[utf8]{inputenc}  
\usepackage[T1]{fontenc}     
\usepackage{lmodern}         
\usepackage{microtype}       
\usepackage[scaled]{helvet} 
\usepackage[english]{babel}  
\usepackage{relsize} 

\usepackage{caption}
\usepackage{subcaption}

\usepackage[normalem]{ulem}
\useunder{\uline}{\ul}{}
\biboptions{sort&compress}   
\usepackage[section]{placeins}
\usepackage{tabularray}
\usepackage{xcolor,hyperref}
\usepackage{tikz}

\begin{document}

\DeclareRobustCommand{\orcidicon}{
	\begin{tikzpicture}
	\draw[lime, fill=lime] (0,0) 
	circle [radius=0.16] 
	node[white] {{\fontfamily{qag}\selectfont \tiny ID}};
	\draw[white, fill=white] (-0.0625,0.095) 
	circle [radius=0.007];
	\end{tikzpicture}
	\hspace{-2mm}
}
\foreach \x in {A, ..., Z}{\expandafter\xdef\csname
orcid\x\endcsname{\noexpand\href{https://orcid.org/\csname orcidauthor\x\endcsname}
			{\noexpand\orcidicon}}
}
\newcommand{\orcidauthorA}{0000-0002-9255-5446} 
\newcommand{\orcidauthorB}{0000-0003-3457-1314}


\title{Deep Learning Innovations for Underwater Waste Detection: An In-Depth Analysis}
\author{Jaskaran Singh Walia\fnref{label1}\orcidA{}%
}
\author{Pavithra L K\fnref{label2}\orcidB{}
}

\author{\\  Department of Computer Science and Engineering, Vellore Institute of Technology, Chennai, India}

\fntext[label1]{karanwalia2k3@gmail.com}
\fntext[label2]{pavithrra.pavi@gmail.com}



\begin{frontmatter}
\begin{abstract}
Addressing the issue of submerged underwater trash is crucial for safeguarding aquatic ecosystems and preserving marine life. While identifying debris present on the surface of water bodies is straightforward, assessing the underwater submerged waste is a challenge due to the image distortions caused by factors such as light refraction, absorption, suspended particles, color shifts, and occlusion. This paper conducts a comprehensive review of state-of-the-art architectures and on the existing datasets to establish a baseline for submerged waste and trash detection. The primary goal remains to establish the benchmark of the object localization techniques to be leveraged by advanced underwater sensors and autonomous underwater vehicles. The ultimate objective is to explore the underwater environment, to identify, and remove underwater debris. The absence of benchmarks (dataset or algorithm) in many researches emphasizes the need for a more robust algorithmic solution. Through this research, we aim to give performance comparative analysis of various underwater trash detection algorithms.
\end{abstract}
\end{frontmatter}



\section{Introduction}  {
In recent years, there has been a significant rise in underwater debris as a result of poor waste management methods, littering, and the expansion of global industries. This waste has caused multiple environmental problems, including impairment of aquatic life and water pollution \cite{COYLE2020100010,Derraik2002ThePO}. When trash is dumped into the water, it not only scatters on the surface but also makes its way down to the epipelagic layer and remains there for years \cite{88df61e7069a431085e274d3c9068466}, polluting the water \cite{https://doi.org/10.48550/arxiv.1803.10813} and harming aquatic animals. Extracting debris from below the surface of the water is difficult and cost inefficient. This raises a requirement for a solution that can work efficiently in a variety of settings at low costs requiring minimal computational needs while delivering high inference speeds.

With the latest advancements in the field of robotics and AI, \cite{10.1007/s11042-020-08976-6,https://doi.org/10.48550/arxiv.1803.10813} it is possible to remove submerged debris with intelligent robots. However, there are challenges associated with the high cost of existing approaches, slow detection speeds, and high computational and memory demands. Several datasets also appear to be highly localized to the particular environment they were collected in, which curtails their capacity to produce a generalized and a robust model.

Autonomous underwater vehicles (AUVs) deployment for trash detection and removal are an essential component of an effective strategy to remove debris from aquatic ecosystems and the initial step for such AUVs consists of detecting the submerged debris, in any specific underwater environment. This is addressed by evaluating the localization using the latest breakthrough strategies in the field of computer vision in order to replace the slower computation algorithms with high memory usage to create a baseline for litter detection. We review a variety of deep learning visual object detection frameworks and use an altered dataset to train and test them, and finally assess their effectiveness using several metrics. In essence, we wish to determine the feasibility and efficacy of using deep learning for real-time visual detection of underwater debris. \\
To be effective in the endeavor to remove plastic and other debris, this paper reviews potential baselines that demonstrate the capability to operate on robotic platforms in near real-time. This requirement is essential for seamless integration into the workflow of waste removal processes.

 }

\section{Review of existing literature}  {
\subsection{Challenges in underwater trash detection}

\begin{figure}[h!]

\centering
  \includegraphics[width=1\linewidth]{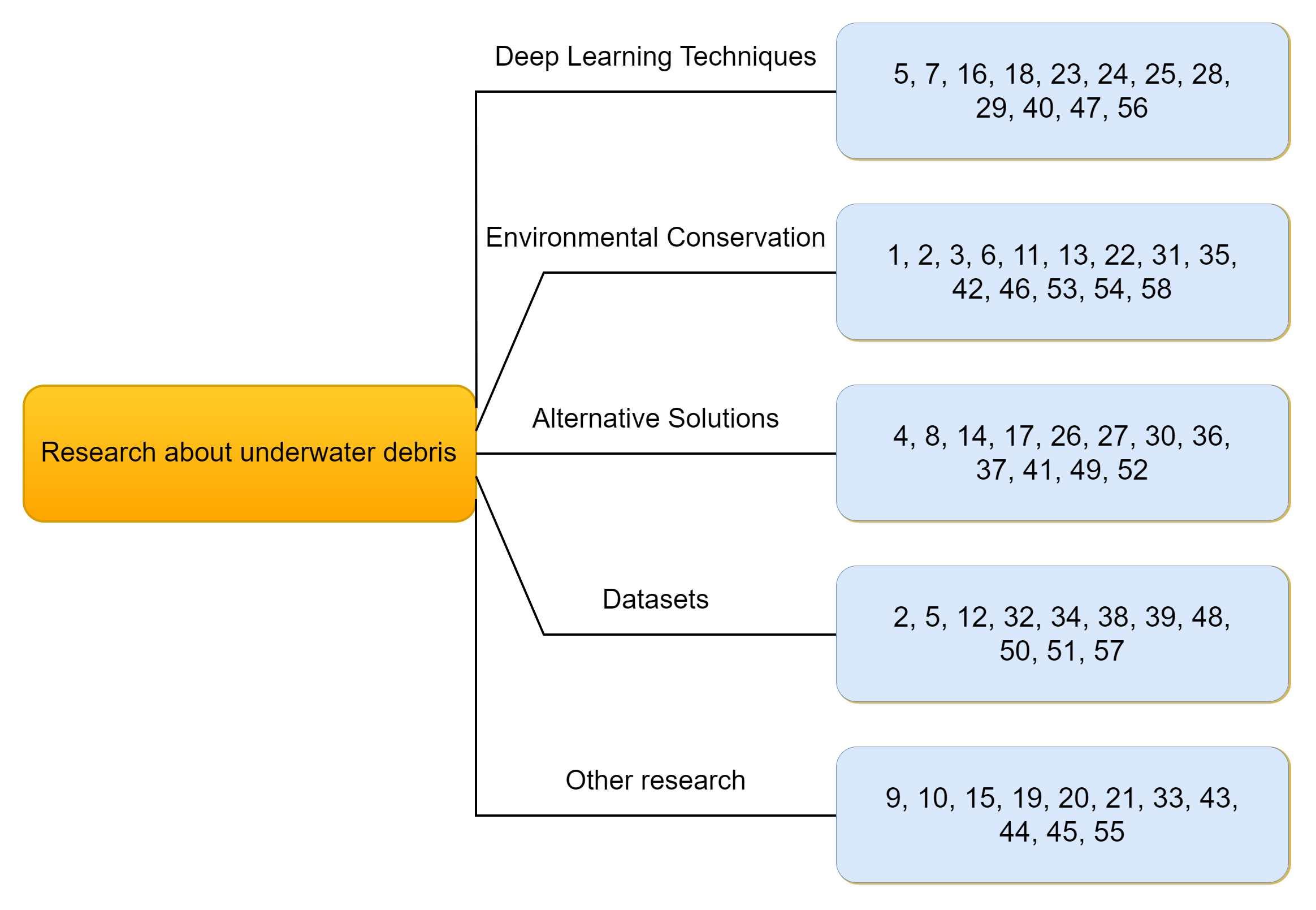}
  \caption{Literature available for the subsequent topics}
  \label{fig:flowchart}

\end{figure}

The detection and management of marine debris are critical to preserving marine ecosystems. The primary challenge lies in the debris' small size relative to the vast ocean, its submerged nature, and the potential for concealment on the sea bottom \cite{028}. Another challenge is the the potential risk to aquatic wildlife\cite{}. A comprehensive review has been done on the various existing literature in this domain explaining the problems faced in harsh marine environments, shown in Figure~\ref{fig:flowchart}

\subsection{Advancements in Detection Technologies}

\subsubsection{Deep Learning and Object Detection}

Advancements in deep learning have propelled the development of object detection algorithms, significantly enhancing marine debris detection capabilities. Figure~\ref{fig:flowchart} \cite{bigpaper} shows a taxonomy of deep learning-based object identification, including detection components, learning algorithms, and applications and benchmarks. This comprehensive approach underpins the ongoing research into autonomous robotic systems\cite{Girdhar2011MARE,temp1,22} for environmental monitoring at both surface and underwater levels to innovative detection\cite{DREVER2018684,tftf,DERRAIK2002842} and documentation\cite{Torrey2009Chapter1T} methods post-natural disasters using ultrasonic\cite{inproceedings,article1,17} and sonar\cite{18} sensors in various aquatic environments\cite{19,20}.

\subsubsection{Technological Innovations in Debris Detection}

Technological innovations extend beyond deep learning approaches that utilize neural networks\cite{018,022,025,026,027} and image processing\cite{029,030,031}, incorporating LIDAR \cite{17}, sonar \cite{18}, and light scattering techniques \cite{19,20} for underwater image enhancement. The operational use of multispectral images by unmanned aerial vehicles for macro-litter mapping and categorization represents another significant leap forward \cite{22}.

\subsection{Deep Learning's Pivotal Role in Underwater Debris Detection}

Deep learning not only facilitates the detection of marine debris but also enhances the quality of underwater imagery. Notable studies employing deep learning for debris detection include the works of ocean pollution detection\cite{018}, identification of floating plastic\cite{022}, deep sea debris detection\cite{025,026}, submerged debris detection\cite{027}, haze removal\cite{029}, localization of trash\cite{030}, as well as a hybrid loss function considering illumination and detection problems \cite{52} and custom networks for detection\cite{031}. These efforts demonstrate the efficacy of deep neural networks in identifying underwater debris, with innovations such as the 'YOLOTrashCan' network emerging as a potent tool for marine debris identification \cite{030,031}.

\subsection{Datasets and Environmental Impact Assessments}

The development of specialized datasets \cite{128,21,23,4524846,tftf} like UIDEF \cite{128}, along with studies on the distribution of deep-sea litter \cite{21} and beach litter segmentation \cite{23}, provides critical resources for training and evaluating detection algorithms. The contributions of research institutions, exemplified by the Monterey Bay Aquarium Research Institute's long-term data collection \cite{4524846} and JAMSTEC's dataset compilation \cite{tftf}, are invaluable to the field. In addition, the TrashCan\cite{trashcan} Dataset has recently emerged as a valuable resource, further enriching the available datasets for training and evaluating detection algorithms.

\subsection{Integrating Environmental Conservation Efforts}

Technological innovations are pivotal for environmental conservation, particularly in addressing marine debris and ecosystem degradation. Clear Blue Sea’s Floating Robot for Eliminating Debris (FRED) initiative \cite{Derraik2002ThePO} deploys autonomous robots equipped with advanced sensors and navigation systems to detect and collect debris in coastal waters. FRED, designed to handle diverse marine environments, is engineered to navigate autonomously, allowing it to efficiently capture floating debris before it reaches the open ocean. The robot's design is modular, enabling it to be adapted for different debris sizes and ensuring versatile functionality.
Similarly, The Rozalia Project employs underwater rovers, notably the ROV (Remotely Operated Vehicle) Hector \cite{88df61e7069a431085e274d3c9068466}, which is specifically designed for oceanic cleanup operations. These rovers are equipped with specialized tools that enable the collection of debris from sensitive marine environments like coral reefs, minimizing disturbances to marine life. The Rozalia Project’s multifaceted approach includes public education and scientific research, emphasizing the importance of maintaining marine ecosystems while exploring innovative solutions for debris removal. \\Both initiatives illustrate the potential for robotics and technology to address marine pollution on a global scale, demonstrating the effectiveness of targeted interventions in preserving marine biodiversity and reducing ecosystem degradation. These projects highlight the transformative potential of technology in preserving marine ecosystems \cite{NOAA,4524846,17,019}, emphasizing the importance of technological innovation in combating marine debris and safeguarding the environment.

\subsection{Exploring Alternative Solutions and Innovations}

The exploration of alternative solutions, such as the NOAA’s monitoring initiatives  \cite{NOAA}, which utilize satellite imagery and advanced data analytics to track and predict debris accumulation patterns, and Howell’s investigation into marine detritus \cite{4524846}, which leverages field research and oceanographic modeling to understand the sources and distribution of debris, signifies a multifaceted approach in understanding and mitigating marine debris. These efforts aim to develop effective strategies for debris management and prevention by providing crucial data on debris hotspots and movement, thus guiding targeted cleanup operations and informing policy decisions.\\These initiatives are instrumental in enriching datasets for underwater object detection algorithms by providing valuable real-world observations. NOAA's monitoring efforts offer comprehensive data on marine environments, aiding in algorithm refinement by enhancing the accuracy of detection systems. Howell's research delves into the composition and distribution of marine debris, offering crucial insights that contribute to algorithm optimization and the overall effectiveness of debris detection methods.

\subsection{Broadening the Scope of Marine Debris Detection Models}
The broad scope of marine debris detection models is evident in the recent contributions to the field, with studies emphasizing diverse approaches. Guo et al. \cite{021} explore unsupervised underwater image enhancement using transformers, significantly improving image clarity for better detection. A comprehensive review of image analysis methods for microorganism counting, including deep learning approaches, which are essential for understanding the impact of marine debris on microscopic life was performed by Li et al. \cite{023}. Another comparative study on different learning algorithms to classify mangrove species using UAV multispectral images was performed by \cite{024}, aiding in the identification of marine debris in coastal ecosystems. Marin et al. \cite{129} investigate deep-feature-based approaches to marine debris classification, revealing the potential of machine learning in accurately categorizing various debris types. Other researchers highlight the importance of accurate classification through a review and case study of marine microdebris\cite{130} and also discuss the use of both remote and in situ devices for assessing marine contaminants\cite{131}, emphasizing the role of advanced technology in environmental monitoring.\\UAV imagery and deep learning are also utilized to assess anthropogenic marine debris in the Maldives\cite{132}, demonstrating the effectiveness of aerial surveys. Vashisht et al.\cite{133} delves into object detection methods and image classification for marine debris, providing essential insights into identifying debris patterns\cite{134}.Salgado-Hernanz et al. \cite{135} underscore recent approaches in remote sensing for marine litter assessment, pointing toward future goals in debris detection. Mittal et al. in his research \cite{27} focussed on underwater image classification through deep learning techniques.\\Additionally, the classification of marine microdebris \cite{130} and the use of remote sensing for marine debris detection \cite{131,132} further expand the methodologies available for marine conservation efforts.

\subsection{Contributions to Underwater Image Classification and Debris Detection}

Contributions to underwater image classification and debris detection are continually evolving, with significant overviews and studies provided by Vashisht \cite{133,134}, along with research in remote sensing \cite{135}, and survey based study of deep learning architectures\cite{27}. These contributions, along with the development of convolutional architectures for sonar images \cite{DBLP:journals/corr/abs-2108-06800} and the Adaptive Lighting Enhancement algorithm \cite{100}, underscore the ongoing advancements in the field.

\subsection{Concluding Remarks on Environmental Implications and Future Directions}

The detection and management of marine debris necessitate a comprehensive and multifaceted approach that incorporates deep learning, technological innovations, and concerted environmental conservation efforts. The body of research, including the operationalization of new detection models and the development of enriched datasets, lays a solid foundation for future advancements. Nonetheless, as the field progresses, it is crucial to consider the environmental implications of these technologies and strive for sustainable and ecologically responsible solutions.
}
\section{Dataset} \label{sec:DS}  {
\subsection {\textbf{Publically available datasets}}  {
Numerous open source underwater image datasets exist on the internet, but a common shortcoming is their high specificity and adaptation to the collection environment, introducing potential bias in the model. Some examples of these existing datasets include TrashCAN 1.0 \cite{trashcan} (both instance and material versions), Trash\textunderscore{}ICRA19\cite{icra19}, JAMSTEC\cite{jam} and \cite{Walia_2023}. Current studies towards automatic waste detection using these datasets are hardly comparable due to the lack of benchmarks and widely accepted standards regarding the used metrics and data \cite{MAJCHROWSKA2022274}. In our previous study\cite{Walia_2023}, we introduced a custom dataset designed to address these challenges and enhance the efficiency of trash localization.

}
\subsection{\textbf{Dataset Comparison}}  {

A comprehensive comparison of all the properties like size, classes, collection location, water-type was performed on existing datasets \cite{https://doi.org/10.48550/arxiv.2105.06808} in order to identify each dataset's limitations and therefore utilizing this information to make the model trained on these datasets more robust.

The diverse range of data prompted an analysis of these datasets which is available in the paper's source repository.\href{https://github.com/karanwxliaa/Underwater-Trash-Detection/tree/main/Custom\%20Models\%20on\%20Public\%20Datasets}{\textsuperscript{2}}.

\begin{table}[h!]
\caption{Comparison of the classes present in various different datasets.} 
\centering

\resizebox{\columnwidth}{!}{%
\begin{tabular}{|c|c|c|c|c|c|c|c|c|}
\hline
{\ul \textbf{Dataset}} &
  \textbf{\begin{tabular}[c]{@{}c@{}}Number of\\ Classes\end{tabular}} &
  {\color[HTML]{2E2E2E} { \textbf{Bio}}} &
  {\color[HTML]{2E2E2E} \textbf{\begin{tabular}[c]{@{}c@{}}Paper\\ and\\ Glass\end{tabular}}} &
  {\color[HTML]{2E2E2E} \textbf{\begin{tabular}[c]{@{}c@{}}Metal waste \\ and\\ Plastic\end{tabular}}} &
  {\color[HTML]{2E2E2E} \textbf{\begin{tabular}[c]{@{}c@{}}Other\end{tabular}}} &
  { \textbf{Unknown}} &
  \textbf{\begin{tabular}[c]{@{}c@{}}Total\\ Images\end{tabular}} &
  \textbf{\begin{tabular}[c]{@{}c@{}}Environmental\\ Conditions\end{tabular}} \\ \hline
Trash ICRA-19 &
  3 &
  1966 &
  0 &
  5051 &
  0 &
  0 &
  6148 &
  \begin{tabular}[c]{@{}c@{}}Normal\\ Underwater\end{tabular} \\ \hline
{\color[HTML]{2E2E2E} TrashCan 1.0} &
  3 &
  {\color[HTML]{2E2E2E} 1009} &
  {\color[HTML]{2E2E2E} 0} &
  {\color[HTML]{2E2E2E} 2163} &
  {\color[HTML]{2E2E2E} 0} &
  180 &
  7212 &
  \begin{tabular}[c]{@{}c@{}}Normal\\ Underwater\end{tabular} \\ \hline
{\color[HTML]{2E2E2E} TrashNet} &
  6 &
  {\color[HTML]{2E2E2E} 0} &
  {\color[HTML]{2E2E2E} 1300} &
  {\color[HTML]{2E2E2E} 0} &
  {\color[HTML]{2E2E2E} 0} &
  0 &
  2527 &
  Clear Background \\ \hline
{\color[HTML]{2E2E2E} Google search} &
  100+ &
  {\color[HTML]{2E2E2E} 92} &
  {\color[HTML]{2E2E2E} 250} &
  {\color[HTML]{2E2E2E} 0} &
  {\color[HTML]{2E2E2E} 0} &
  366 &
  1320 &
  Varying \\ \hline
{\color[HTML]{2E2E2E} Extended TACO} &
  28 &
  {\color[HTML]{2E2E2E} 70} &
  {\color[HTML]{2E2E2E} 1190} &
  {\color[HTML]{2E2E2E} 6060} &
  {\color[HTML]{2E2E2E} 2800} &
  150 &
  1500 &
  Wild Waste \\ \hline
{\color[HTML]{2E2E2E} Drinking waste} &
  4 &
  {\color[HTML]{2E2E2E} 0} &
  {\color[HTML]{2E2E2E} 1160} &
  {\color[HTML]{2E2E2E} 3600} &
  {\color[HTML]{2E2E2E} 0} &
  0 &
  9640 &
  \begin{tabular}[c]{@{}c@{}}Clear Conditions,\\ Cans \& Bottles\end{tabular} \\ \hline
Classify-waste &
  3 &
  160 &
  3900 &
  9660 &
  2800 &
  520 &
  27500 &
  Varying \\ \hline
\end{tabular}%
}

\end{table}

The used dataset \cite{Walia_2023} was constructed using three major steps: data collection, manual annotation, and data pre-processing using rotation, random cropping, horizontal flips, expansion, and color jittering \cite{106, 146}. This augmented approach to data has demonstrated notable enhancements in detection accuracy. The final dataset contained 10,000 images of size 416x416 mapped into three classes: (i) Underwater Trash, (ii) Rover, and (iii) Biological life.
We compare the different existing datasets relevant to this use case based on various factors, such as the number of classes, total images, environmental conditions, and the different classes, from Table 1.
}
}

\section{Methodology}  {
With recent advancements in computer vision technologies and algorithms, this section introduces cutting-edge identification and categorization models, accompanied by established standards for the utilized dataset. The training metrics are then statistically evaluated for each model providing insights into their efficiencies and potential for real-world application.

 \subsection{\textbf{Object Localization }}  {
The object detection architectures were carefully selected from the latest, highly coherent, and promising set of networks currently available. Each architecture has its own unique strengths and weaknesses, offering varying degrees of reliability, execution pace, and other metrics which we identify in this literature. \\
In this paper we also leverage several neural network architectures considered to be state-of-the-art models, including YOLOv9, YOLOv8 nano, YOLO v7, YOLO v6s, YOLO v5s, and YOLO v4 Darknet, utilizing their respective source repositories \cite{yolo}. In addition, custom training was performed on region-based Mask-RCNN and FasterRCNN architectures with Inception v2 \cite{inception}.
 }
 \subsection{\textbf{Object Segmentation}}  {
Building on the advancements in computer vision, this section focuses on cutting-edge object segmentation methodologies.  Real-time instance segmentation, exemplified by YOLACT and Mask R-CNN models trained on underwater litter data, offers promising prospects for litter detection, with Mask R-CNN achieving a mean average precision (mAP) of 0.377 and YOLACT providing faster detection speeds\cite{seg1}. AquaSAM\cite{seg2} enhances underwater segmentation, particularly in tasks like coral reef segmentation, showcasing a notable 7.13\% improvement in Dice Similarity Coefficient. A novel deep learning-based method for trash trap defect detection\cite{seg3} demonstrates significant enhancements in reliability, with error rates below 15\% and accuracy rates exceeding 85\%. Furthermore, leveraging Cycle Consistent Generative Adversarial Networks (GANs), enhancing unpaired underwater images exhibits superior performance over traditional techniques\cite{seg4}, showing potential for various marine applications such as trash identification and coral reef inspection.State-of-the-art segmentation models are considered, aligning with the latest advancements in technology. The chosen models adhere to predefined standards for the dataset, ensuring a rigorous evaluation of the segmentation performance. Each segmentation architecture brings its distinctive features, strengths, and weaknesses, contributing to a comprehensive understanding of their efficacy.
 }
 \subsection{\textbf{Graphics Processing Unit}}  {
In this study, we utilized a Tesla T4 GPU with a total memory capacity of 16GB, comprising 40 Streaming Multiprocessors (SMs) and a shared 6MB L2 cache across all SMs. Released in 2018, this GPU is equipped with 2560 CUDA cores. 
 }

\subsection{\textbf{Architectures}}{
This section provides an overview of latest state-of-the-art models employed in the research and presents the outcomes achieved during experimentation.\\
Presently, object detection frameworks rooted in deep learning can primarily be categorized into two main groups: (i) two-stage detectors, exemplified by Region-based CNN (R-CNN) \cite{rcnn2} and its variations \cite{38, 34, 39}, and (ii) one-stage detectors, such as YOLO \cite{40} and its variations \cite{41, 42}. Two-stage detectors initially utilize a proposal generator to produce a sparse set of proposals and then extract features from each proposal. This is followed by region classifiers that predict the category of the proposed region. In contrast, one-stage detectors directly predict the category of objects at each location on the feature maps without the additional region classification step. While two-stage detectors commonly achieve superior detection performance and report state-of-the-art results on public benchmarks, one-stage detectors are notably more time-efficient and exhibit greater suitability for real-time object detection\cite{bigpaper}.

\subsubsection{\textbf{You Only Look Once (YOLO)}}{
YOLO, an acronym for You Only Look Once, stands out as a prominent and sought-after object identification method. It excels in the real-time identification of multiple items within a coexisting image or video. Unlike traditional methods, YOLO employs a single neural network to predict bounding boxes and class probabilities directly from the entire image in a single analysis. This approach enhances efficiency and precision, making YOLO a superior choice compared to other existing object detection and identification systems. A comparative analysis in Table 2 underscores YOLO's prowess, showcasing its superior performance in terms of model size, mean average precision (mAP), speed, and computational parameters across different variants like YOLOv8n, YOLOv8x, and YOLOv5s.

\begin{figure}[h!]

\centering
  \includegraphics[width=1\linewidth]{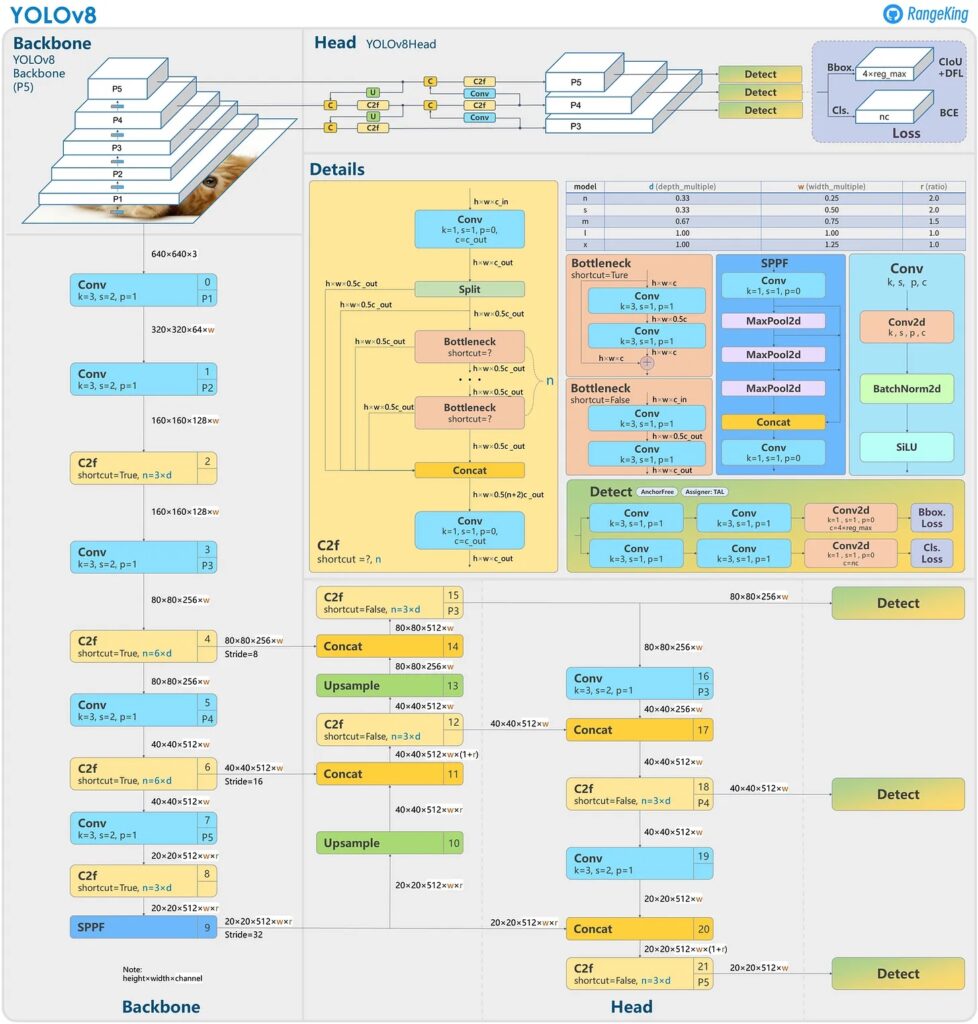}
  \caption{Architecture of the YOLOv8 model \cite{yolov8}}

\end{figure}
\begin{itemize}
  \item \textbf{YOLOv8x \cite{yolov8} (XLarge)}: Building upon the innovations of its predecessors, YOLOv8x represents an advanced variant of YOLOv8, with a primary focus on achieving heightened accuracy, particularly for small and intricate objects. This improvement is realized through an augmentation in model size, achieved by enhancing both depth and width. The increased dimensions enable the model to capture more intricate visual patterns, resulting in superior detection performance, especially in challenging scenarios. YOLOv8x outperforms YOLOv8n, making it particularly well-suited for applications demanding exceptional precision. Its larger model size enhances object understanding and localization, thereby significantly improving overall object detection capabilities.
  \item \textbf{YOLOv8n \cite{yolov8} (Nano)}: The YOLOv8n model presents a swifter and more precise unified framework for training models, with its nano version standing out as the fastest and most compact among the available models. However, the heightened speed of the model comes at the cost of detailed prediction accuracy. Despite this trade-off, the nano version proves to be the most suitable for real-time deployment owing to its rapid detection rates and lower memory requirements. Notably, it introduces anchor-free detection, negating the necessity for anchor box adjustments and streamlining the detection process.\\
  Moreover, YOLOv8n incorporates novel convolutions by substituting the initial 6x6 convolution in the stem with a more efficient 3x3 convolution. Further modifications are made to the concatenation of features in the neck. An advanced training routine is employed, encompassing mosaic augmentation to introduce variations during the training process.
  \item \textbf{YOLOv7 \cite{https://doi.org/10.48550/arxiv.2207.02696} (tiny/W6)}: The YOLOv7 algorithm has markedly outperformed its earlier versions in terms of both speed and precision. Notably, it requires less sophisticated hardware compared to standard conventional architectures. YOLOv7 demonstrates an impressive ability to be trained effectively on smaller datasets without pre-learned weights within a shorter timeframe. However, it is important to highlight that, despite these strengths, YOLOv7 falls short of achieving superior accuracy and inference speeds when compared to the latest YOLOv8 models.
  \item \textbf{YOLOv5s \cite{glenn_jocher_2022_7347926} and YOLOv6s \cite{https://doi.org/10.48550/arxiv.1512.03385} (Small)}: They exhibit relatively consistent performances and results, making them suitable for benchmarking. However, they fall short when compared to some of the most widely used stable models for object identification and segmentation
\end{itemize}
}
\begin{table}
\caption{Table comparing computation values and speed of the best performing models.} 

\centering
\begin{tblr}{
  width = \linewidth,
  colspec = {Q[133]Q[108]Q[113]Q[158]Q[212]Q[110]Q[100]},
  hlines,
  vlines,
}
\textbf{Model} & {\textbf{size}\\\textbf{(pixels)}} & {\textbf{mAP val}\\\textbf{50-95}} & {\textbf{Speed}\\\textbf{CPU ONNX}\\\textbf{(ms)}} & {\textbf{Speed}\\\textbf{A100 TensorRT}\\\textbf{(ms)}} & {\textbf{params}\\\textbf{(M)}} & {\textbf{FLOPs}\\\textbf{(B)}} \\
YOLOv8n        & 640                                & 37.3                              & 80.4                                               & 0.99                                                    & 3.2                             & 8.7                            \\
YOLOv8x        & 640                                & 53.9                              & 479.1                                              & 3.53                                                    & 68.2                            & 257.8                          \\
YOLOv5s        & 640                                & 56.8                              & 98                                                 & 6.4                                                     & 7.2                             & 16.5                           
\end{tblr}

\end{table}

\subsubsection{\textbf{R-CNN}}{
R-CNN\cite{rcnn2}, introduced by Girshick et al. in 2014, is a groundbreaking two-stage object detection system. Compared to previous state-of-the-art methods like SegDPM \cite{44}, which achieved a mean Average Precision (mAP) of 40.4
\% on Pascal VOC2010, R-CNN showed a significant improvement with 53.7\% mAP. The R-CNN pipeline is divided into three main components: proposal generation, feature extraction, and the region classification.\\R-CNN generates a sparse collection of proposals (about 2,000 for each image) using Selective Search \cite{45}. This technique is intended to filter away regions that can be easily identified as background. A deep convolutional neural network then crops and resizes each proposal into a fixed-size region before encoding it into a feature vector (for example, 4,096 dimensions). Following that, a one-vs-all SVM classifier is used for region classification. Finally, bounding box regressors are trained with the extracted characteristics as input, with the goal of fine-tuning the original suggestions to better capture the objects.\\
Inspired by spatial pyramid matching (SPM) \cite{46}, He et al. proposed SPP-net \cite{47} to boost R-CNN's speed and learn more distinct features. Unlike the traditional method of clipping proposed regions and passing them separately into a CNN model, SPP-net creates the feature map from the full image using a deep convolutional network. It then extracts fixed-length feature vectors from the feature map using a Spatial Pyramid Pooling (SPP) layer.

\subsubsection{\textbf{Fast R-CNN}}{
Fast R-CNN \cite{38} is a multi-task learning detector that outperforms SPP-net while resolving its shortcomings. Similar to SPP-Net, Fast R-CNN generates a feature map for the full picture and extracts fixed-length region features from it. However, unlike SPP-net, Fast R-CNN uses a ROI (Region of Interest) Pooling layer to retrieve region features.
}
\subsubsection{\textbf{Faster R-CNN}}{

Despite significant advancements in learning detectors, the proposal generation step remained reliant on conventional approaches such as Selective Search \cite{45} or Edge Boxes \cite{48}, which were grounded in low-level visual cues and lacked the capacity to learn in a data-driven manner. To tackle this limitation, faster R-CNN, an improvement over R-CNN\cite{rcnn2}, which utilized a Region Proposal Network (RPN) \cite{57,58,59} was introduced, enhancing the creation of an end-to-end trainable object detection network \cite{NIPS2015_14bfa6bb}. The RPN generates region proposals using the last convolutional feature map, followed by fully connected layers for final detection. Although many implementations typically employ VGG-16 as a standard for feature extraction, we chose to use Inception v2, which has demonstrated superior object identification performance in conventional datasets \cite{tensorflowmodelgarden2020, simonyan2014deep}. However, it may misidentify patches in the background as objects due to its limited ability to consider large contexts.\\
When comparing generalization errors, YOLO algorithms surpass R-CNN by producing lower generalization errors. In comparison to the YOLO algorithms, Faster R-CNN exhibits slightly fewer localization errors but at the expense of higher computational needs and slower speed.
}

\subsubsection{\textbf{Mask RCNN}}  {
In order to enhance the overall flexibility of the process, He et al. \cite{3} introduced Mask R-CNN, a model capable of predicting bounding boxes and segmentation masks at the same time, yielding optimal results. Huang et al. \cite{600} developed Mask Scoring R-CNN, a system that prioritizes mask quality awareness. This framework learns the quality of predicted masks and uses calibration approaches to correct the misalignment between mask quality and mask confidence score.
This has also emerged as a faster variant by combining ResNet with a Feature Pyramid Network (FPN) the combination of which creates a feature extraction network capable of generating a multi-layer convolutional feature map from input images \cite{https://doi.org/10.48550/arxiv.1512.03385}. The Region Proposal Network (RPN) is then employed to generate object region proposals, while the detection head and mask head are utilized for object detection and instance segmentation. The algorithm is structured into two main components: detection and instance segmentation.\cite{3,600}\\
Despite its effectiveness, Mask R-CNN can face challenges in complex maritime environments. External variables such as distracting objects and low image resolution may impact the algorithm's performance by making it difficult to extract object features, leading to reduced attention toward objects during detection.
}
}

\subsection{\textbf{Evaluation metrics}}  {
After model training, we utilize distinct evaluation and validation datasets, separate from the training dataset, to assess the accuracy of the architecture. The model accurately classifies trash by creating bounding boxes with a confidence score of 0.50 or higher. The amount of true positive bounding boxes containing marine plastic debris and the accuracy with which true negatives are detected are used to evaluate the model's performance.

These performance metrics were used to evaluate each model's performance, pros and cons:
\begin{itemize}
  \item \textbf{True positive and True negative values:} These values are calculated by counting the number of bounding boxes correctly predicted as trash (true positives) and the number of bounding boxes correctly predicted as not trash (true negatives).

  \item \textbf{Precision and Recall:} Evaluates if the model foreseen trash located in the input image in which precision reflects the model's ability to correctly predict plastic waste, while recall reflects the model's ability to identify all instances of plastic waste in the input image. Precision is calculated as the ratio of true positives to the sum of true positives and false positives, while recall is calculated as the ratio of true positives to the sum of true positives and false negatives\cite{pr}.

\begin{equation}
Recall = \frac{TP}{TP+FN}
\label{eq:recall}
\end{equation}

\begin{equation}
Precision = \frac{TP}{TP+FP}
\label{eq:precision}
\end{equation}

  These equations, \ref{eq:recall} for recall and \ref{eq:precision} for precision, illustrate how the performance of the model is quantitatively determined in terms of its ability to identify plastic waste accurately.

  \item \textbf{Mean Average Precision:} This metric measures the accuracy of the model in identifying underwater trash waste across a set of input images. The mAP is calculated by building a precision-recall curve using the Intersection over Union (IoU) formula to compare the predicted and ground truth bounding boxes \cite{map}. The precision-recall curve is then integrated to obtain the mAP. After collecting the accurate and inaccurate data, we utilize the "Intersection over Union (IoU)" formula to give rise to a precision-recall curve: 

\begin{equation}
 IOU = \frac{B Box_{pred}  \cap B Box_{GroundTruth}}{B Box_{pred}  \cup B Box_{GroundTruth}}
\label{eq:iou}
\end{equation}

Where \(B Box_{Predicted}\) and \(B Box_{Groundtruth}\) represent the expected areas under the curve for ground truth and predicted bounding boxes, respectively. In order to maximize accuracy, it is necessary to set a high threshold for confidence and IoU, ensuring that correct predictions exceed this threshold. Subsequently, The average precision (mAP) can be calculated by integrating the precision-recall curve.\cite{10.1007/978-3-642-40994-3_29}:

\begin{equation}
\mathlarger{mAP = \int_{0}^{1} p(x) dx}
\label{eq:map}
\end{equation}

 \item \textbf{Processing speeds:} This metric denotes the average time taken by the model to perform pre-processing, inference and post-process for the detection and identification of underwater trash in an input image, measured in milliseconds per image.
 \begin{enumerate}
 \item \textbf{YOLOv8n} gave us an average speed of 1.9ms for pre-process, 73.5ms for inference and 0.95ms post-process per image at shape (1, 3, 416, 416)
 \item \textbf{YOLOv8x} gave us an average speed of 0.3ms for pre-process, 16.6ms for inference and 1.4ms post-process per image at shape (1, 3, 416, 416)
\item \textbf{YOLOv5s} gave us an average speed of 0.5ms for pre-process, 12.0ms for inference with 1.2ms post-process per image at shape (1, 3, 640, 640)

 \end{enumerate}

\end{itemize}
}
}

\section{Results}{
\subsection{\textbf{Evaluating results}}{
A comprehensive analysis of individual components for the best performing frameworks was conducted, involving thorough examinations of trash data publicly available in various scenarios, including clean water, natural and man-made water bodies such as lakes and ponds, surroundings, and ocean beds was performed. The utilized architectures also demonstrate a relatively high F1 score and a high mean-average precision (mAP) score relative to their inference speeds. Table 3 shows the results of an in-depth evaluation comparing multiple architectural networks.
\begin{table}[h!]
\caption{Performance comparison for the various architectures used}
\centering
\resizebox{\columnwidth}{!}{%
\begin{tabular}{|l|l|l|l|l|l|l|l|}
\hline
{\ul \textbf{Network}} &
  {\color[HTML]{2E2E2E} {\ul \textbf{mAP}}} &
  {\color[HTML]{2E2E2E} {\ul \textbf{Precision}}} &
  {\color[HTML]{2E2E2E} {\ul \textbf{Recall}}} &
  {\color[HTML]{2E2E2E} {\ul \textbf{Box-Loss}}} &
  {\color[HTML]{2E2E2E} {\ul \textbf{Cls-Loss}}} &
  {\color[HTML]{2E2E2E} {\ul \textbf{Obj-Loss}}} &
  {\ul \textbf{Epoch}} \\ \hline
{\color[HTML]{2E2E2E} YOLOv8n} &
  {\color[HTML]{2E2E2E} 0.96} &
  {\color[HTML]{2E2E2E} 0.94} &
  {\color[HTML]{2E2E2E} 0.92} &
  {\color[HTML]{2E2E2E} 0.022} &
  {\color[HTML]{2E2E2E} 0.0005} &
  {\color[HTML]{2E2E2E} 0.011} &
  170 \\ \hline
{\color[HTML]{2E2E2E} YOLOv7} &
  {\color[HTML]{2E2E2E} 0.96} &
  {\color[HTML]{2E2E2E} 0.96} &
  {\color[HTML]{2E2E2E} 0.93} &
  {\color[HTML]{2E2E2E} 0.019} &
  {\color[HTML]{2E2E2E} 0.0005} &
  {\color[HTML]{2E2E2E} 0.008} &
  120 \\ \hline
{\color[HTML]{2E2E2E} YOLOv6s} &
  {\color[HTML]{2E2E2E} 0.90} &
  {\color[HTML]{2E2E2E} 0.94} &
  {\color[HTML]{2E2E2E} 0.92} &
  {\color[HTML]{2E2E2E} 0.08} &
  {\color[HTML]{2E2E2E} 0.07} &
  {\color[HTML]{2E2E2E} 0.45} &
  110 \\ \hline
{\color[HTML]{2E2E2E} YOLOv5s} &
  {\color[HTML]{2E2E2E} 0.96} &
  {\color[HTML]{2E2E2E} 0.95} &
  {\color[HTML]{2E2E2E} 0.93} &
  {\color[HTML]{2E2E2E} 0.02} &
  {\color[HTML]{2E2E2E} 0.00} &
  {\color[HTML]{2E2E2E} 0.00} &
  180 \\ \hline
{\color[HTML]{2E2E2E} Faster R-CNN} &
  {\color[HTML]{2E2E2E} 0.81} &
  {\color[HTML]{2E2E2E} 0.88} &
  {\color[HTML]{2E2E2E} 0.85} &
  {\color[HTML]{2E2E2E} 0.08} &
  {\color[HTML]{2E2E2E} 0.03} &
  {\color[HTML]{2E2E2E} 0.02} &
  100 \\ \hline
Mask R-CNN &
  0.83 &
  0.85 &
  0.87 &
  0.09&
  0.01 &
  0.08&
  100 \\ \hline
\end{tabular}%
}

\end{table}
The observed trade-offs indicate that the outcomes derived in this research provide a more accurate representation of performance over time across a broader spectrum of water conditions, allowing for an in-depth analysis of the various model's object localization performance in this domain. From table 3 we conclude that both YOLOv8-nano and YOLOv8-XLarge successfully achieve real-time localization of epipelagic detritus, demonstrating robust metrics. Furthermore, they exhibit a notably higher F1 score compared to the other tested algorithms.

\begin{figure}[h!]

  \includegraphics[scale=.26]{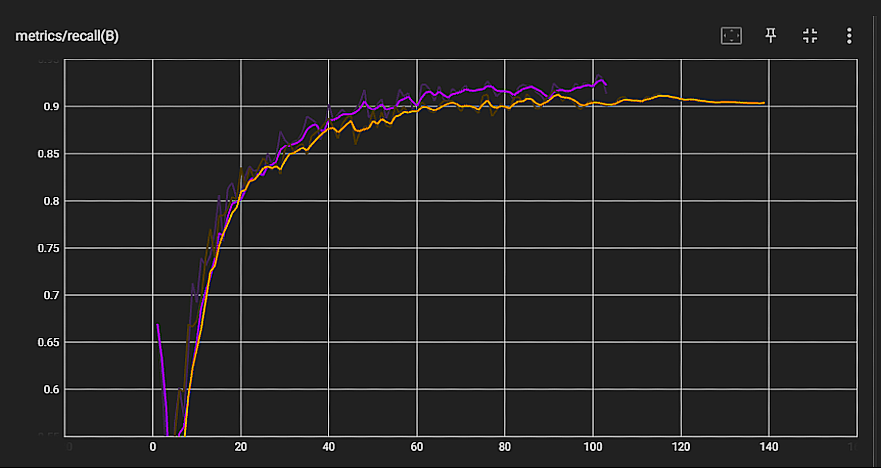}
  \includegraphics[scale=.26]{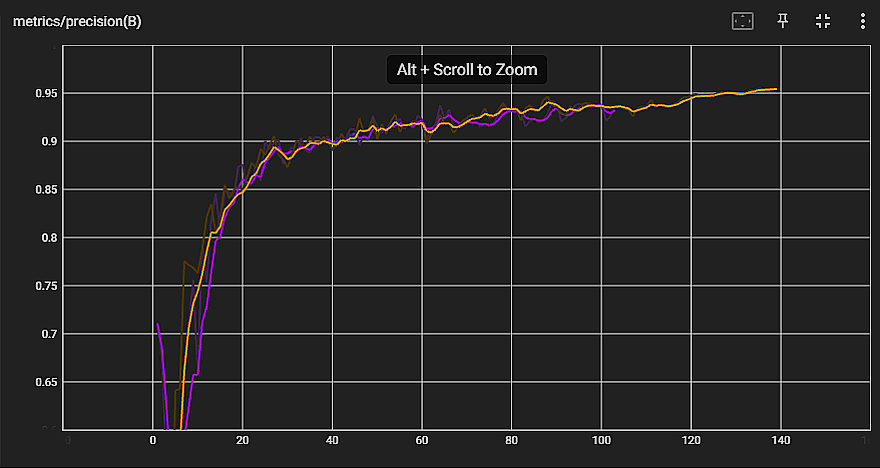}

\caption{Precision and Recall values (Y-axis) over 140 epochs (X-axis) for YOLOv8n and YOLOv8x with colors Purple and Yellow respectively}

\end{figure}

\begin{figure}[]
  \includegraphics[scale=.5]{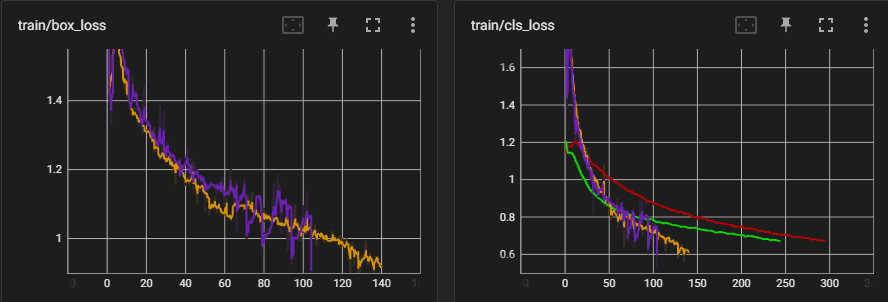}
  \centering
\caption{Loss values (Y-axis) over 140 epochs (X-axis) for YOLO v5s,v7,v8n,v8x : Green, Red, Violet and Yellow}

\end{figure}

Additionally, advanced algorithms within the same family, including YOLOv5, v7, YOLO-NAS, and YOLO-SAM, were also evaluated, yielding similar outcomes. The decision was made to opt for the lighter models like tiny, small and nano networks due to their quickest evaluation performance, fewer parameters, and lower processing power requirements. These algorithms also outperformed classic Mask R-CNN and and Faster R-CNN algorithms, showcasing significantly higher speeds \& better F1 scores.
\begin{figure}[h!]
\centering
  \includegraphics[width=1\linewidth]{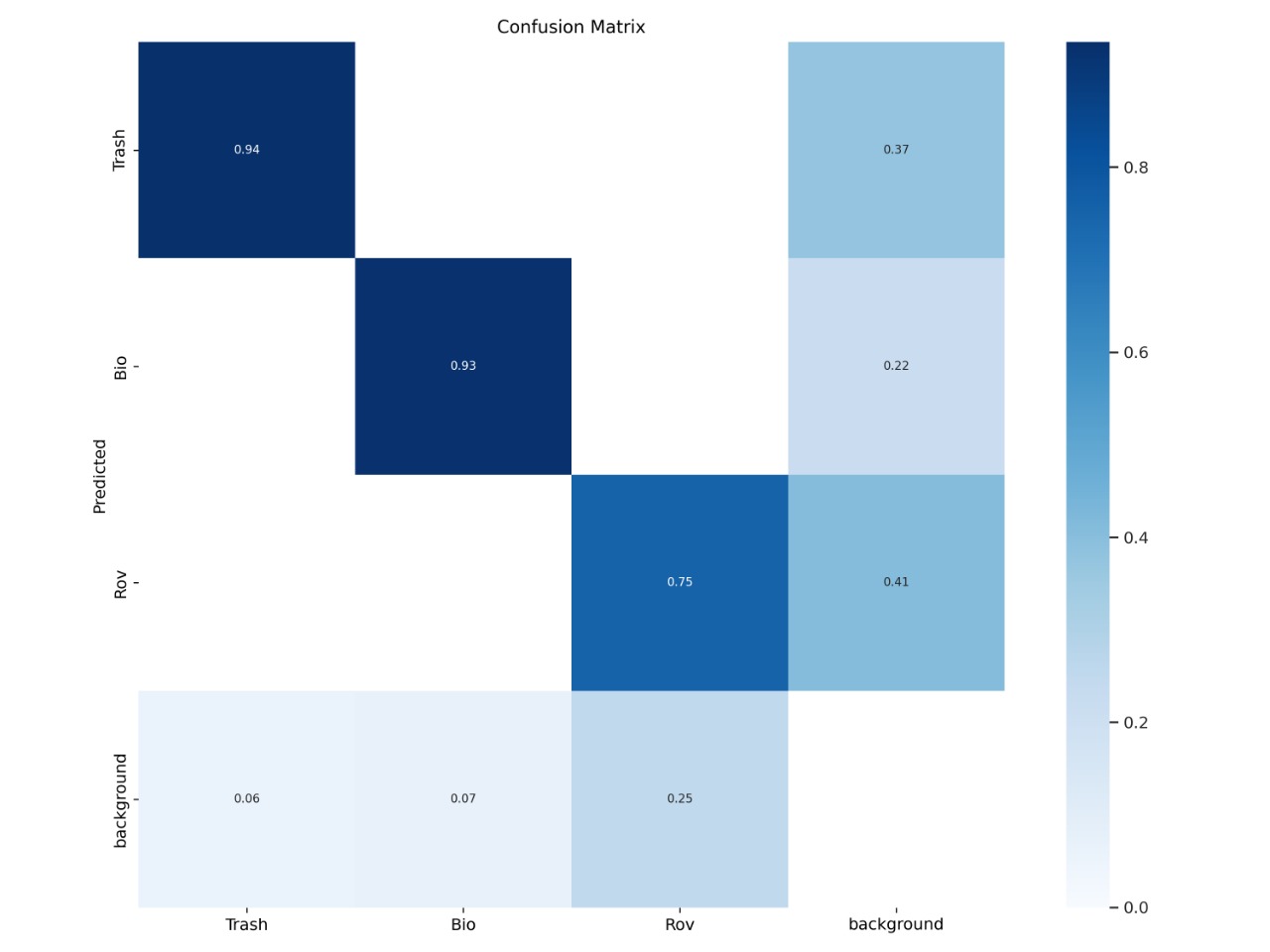}
  \caption{Confusion matrix for the best performing YOLO model}
\end{figure}
The model's performance in real-world deployments mirrored the results obtained in controlled evaluations, with only minor deviations between the outcomes presented in Table 4 and those observed in a real-time environment. 
}
\subsection{\textbf{Visualising results}}{
As part of its image processing, the network captures two important data points in arrays:\\ \textbf{Scores}, which reflect the degree of certainty for the predicted boxes, and the total number of \textbf{Detections} made on the image.\\
In order to render the bounding boxes on the images, the network converts the normalized coordinates to image coordinates, utilizing a specific equation.
\[ 
ImgCord_{k} = BoxScore_{i}^{j} \ast Width 
\]
In the given equation, $k$ represents all corners (right, left, top, \& down), $j$ belongs to the set ${0, 1, 2, 3}$, $i$ denotes the box index, and $Width$ indicates the image width. During the evaluation of test images, the coordinates of the image were utilized to visualize the obtained results of the bounding boxes.
The test dataset, which was deliberately designed to provide a representation of unseen test data new to the model to test the model's performance on new data was used to obtain the results presented in Figures 6,7 and 8. Details on the dataset\cite{Walia_2023} benchmarks can be found in the \hyperref[sec:DS]{Dataset Section $^3$}.
\begin{figure}[h!]
\centering
\begin{subfigure}{.5\textwidth}
  \centering
  \includegraphics[width=1\linewidth]{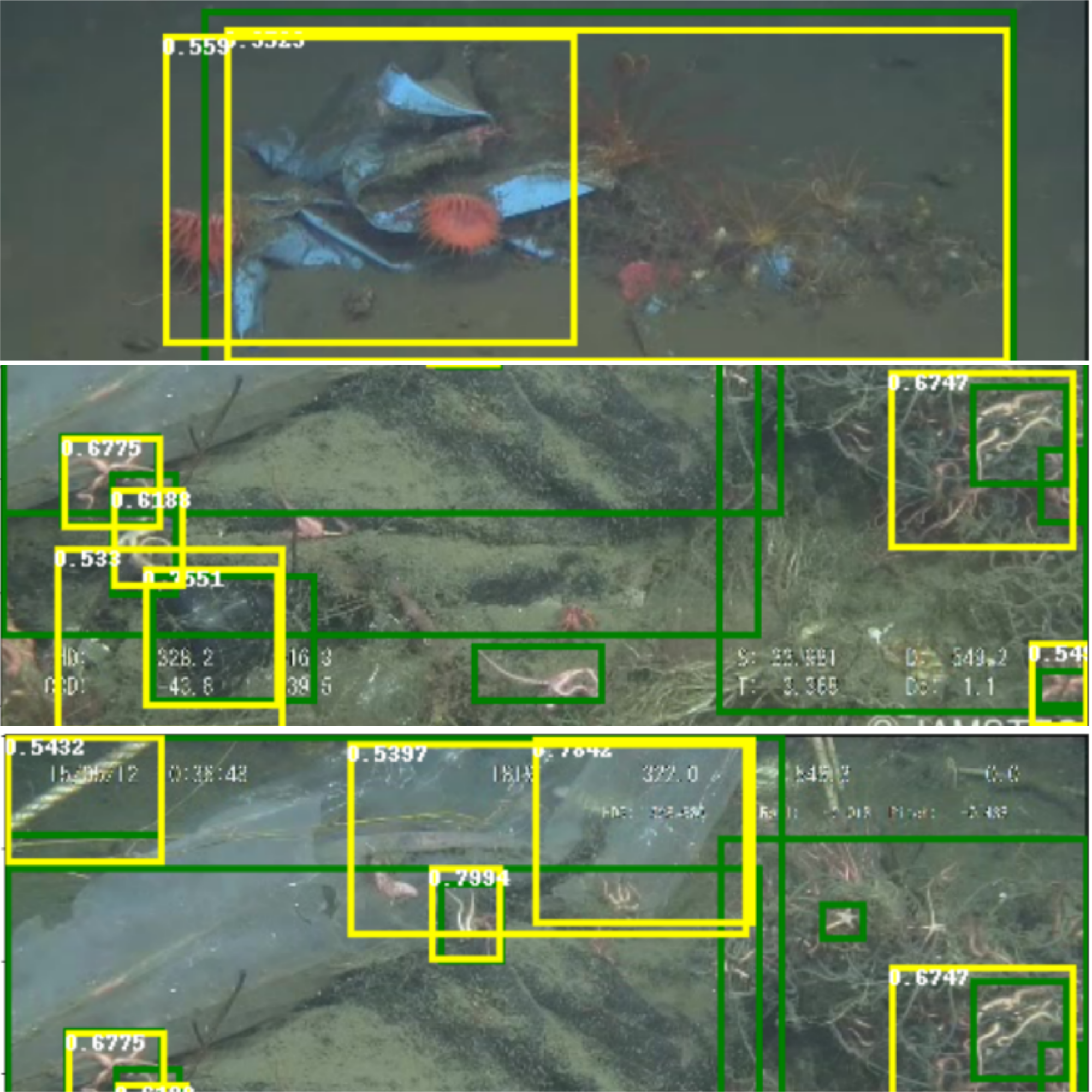}
  \caption{Batch results for Faster-RCNN}
  \label{fig:sub1}
\end{subfigure}%
\begin{subfigure}{.5\textwidth}
  \centering
  \includegraphics[width=1\linewidth]{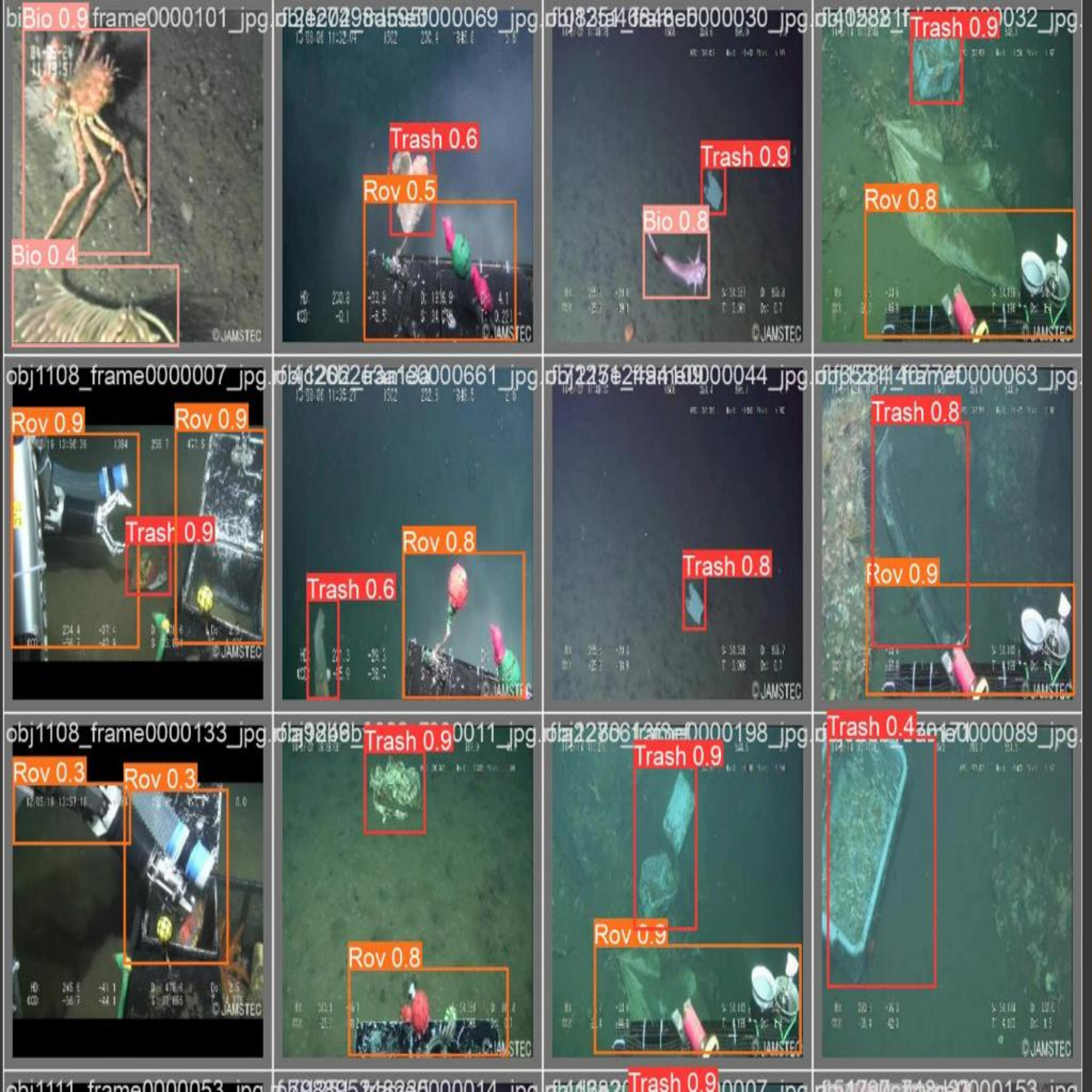}
  \caption{Batch results for YOLO models}
  \label{fig:sub2}
\end{subfigure}
\caption{Batch comparison between the detection outputs of RCNN \& YOLO models.}
\label{fig:sub3}
\end{figure}
\\In addition to analyzing the models using our test set, we also subjected each network to two additional videos, as illustrated in the Figure 7 One of the videos shown in Figure  was a viral footage of ocean waste captured in Bali \cite{yt}, which presented a unique and previously unseen visual context compared to our training set. Despite the significant challenges posed by this video to all our detectors, these models were able to provide promising detection. 
\begin{figure}[h]
\centering
  \includegraphics[width=0.8\linewidth]{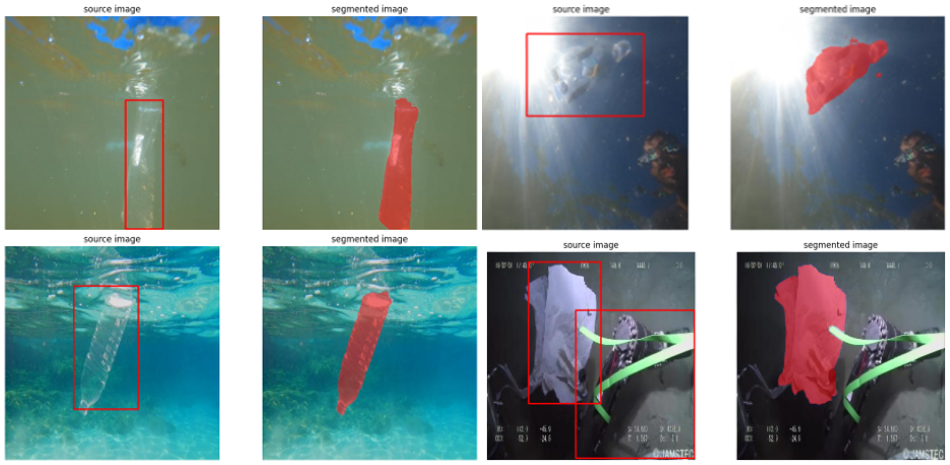}
  \caption{Real-time segmentation results for YOLOv8 models}
\end{figure}
The above figure depicts real-time segmentation results utilizing the proposed model which can further be used for quantization of trash in a particular environment.
\begin{figure}[h]
\centering
  \includegraphics[width=0.8\linewidth]{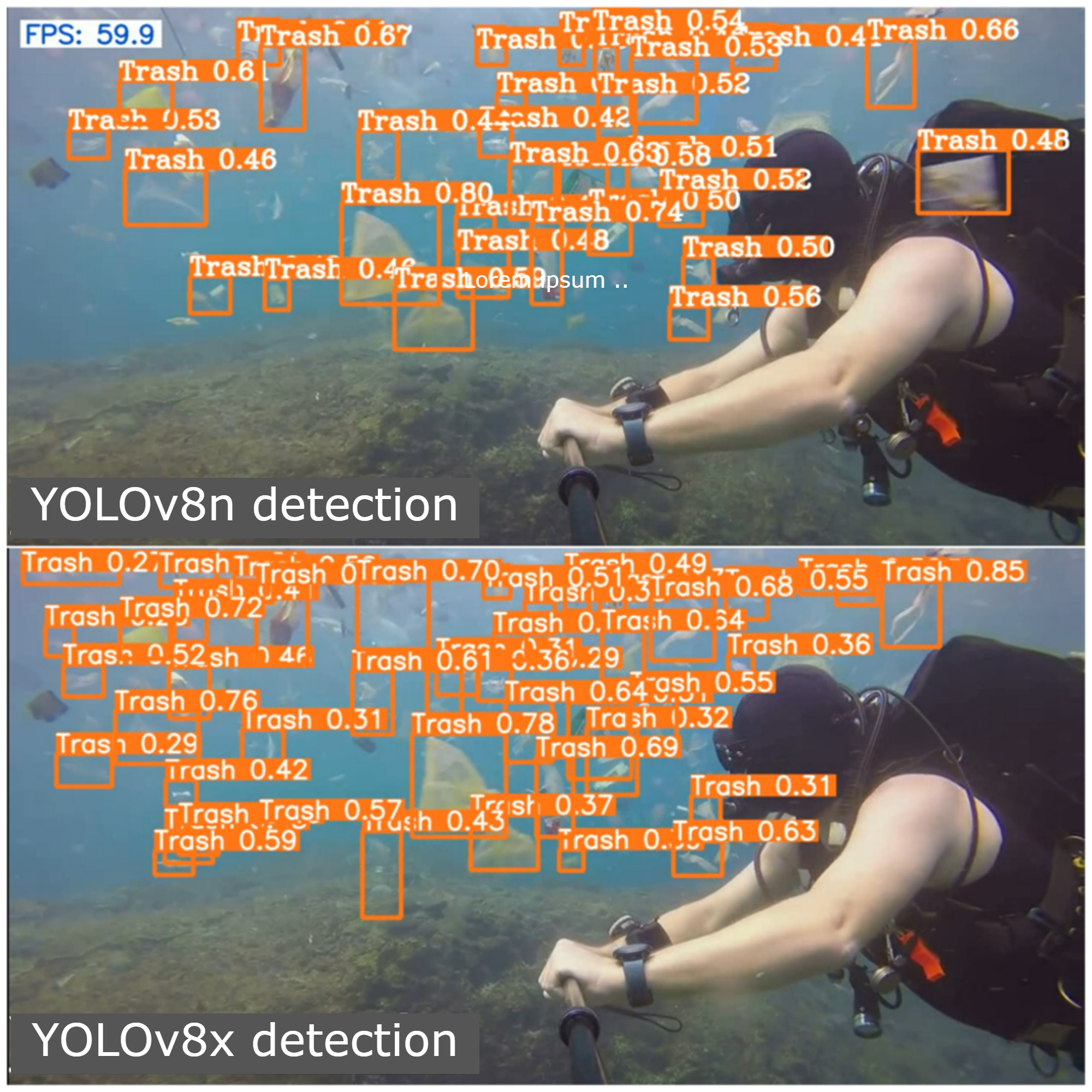}
\caption{Performance on new, unseen test data using the best tradeoff model}
\end{figure}
}
}

\section{Conclusion}{
In conclusion, this research evaluated cutting-edge computer vision architectures for the localization and quantification of underwater submerged trash across various water layers, including the epipelagic and mesopelagic zones, and extending to other local water bodies. Leveraging a carefully curated dataset and employing state-of-the-art architectures such as YOLO, our primary objective of reviewing the feasibility of localizing submerged trash underwater, in real-time inference speeds for trash eradication was assessed.\\
The rapid inference speeds observed in this study attest to the viability of maintaining an impressive level of performance which underscores the potential for utilizing Autonomous Underwater Vehicles (AUVs) for the automatic categorization of diverse underwater objects, as well as the sorting and collection of trash in locations that are challenging for humans to access due to factors such as high pressure as well as other environmental conditions.\\
These findings hold promise for the automation of trash collection and eradication in harsh underwater aquatic environments, showcasing the potential of advanced deep-learning methods in addressing the challenges of underwater debris management. Moreover, this research serves as a crucial foundation and standard for future endeavors that seek to identify and categorize underwater debris, paving the way for advancements in the field.
}

\bibliographystyle{elsarticle-num}

\section*{References}

\bibliography{main}

\end{document}